\documentclass{article}

\usepackage{graphicx}
\usepackage{latexsym}

\newtheorem{theorem}{Theorem}

\newenvironment{proof}{\noindent{\bf Proof.}}{\mbox{}\hfill$\Box$\\ \\}

\newcommand\prob{{\rm prob}}
\newcommand\lab{{\rm label}}
\newcommand\prt{{\rm partition}}

\title{Postponing Branching Decisions}

\author{Willem Jan van Hoeve\footnote{CWI, 
P.O. Box 94079, 1090 GB Amsterdam, The Netherlands} 
\mbox{} and Michela Milano\footnote{DEIS, University of Bologna, 
Viale Risorgimento 2, 40136 Bologna, Italy}}

\date{July 16, 2004}

\begin{document}

\maketitle

\begin{abstract}
Solution techniques for Constraint Satisfaction and Optimisation Problems
often make use of backtrack search methods, exploiting variable and value
ordering heuristics. In this paper, we propose and analyse a very simple
method to apply in case the value ordering heuristic produces ties: {\bf
postponing the branching decision}. To this end, we group together values
in a tie, branch on this sub-domain, and defer the decision among them to
lower levels of the search tree. We show theoretically and experimentally
that this simple modification can dramatically improve the efficiency of
the search strategy. Although in practise similar methods may have been
applied already, to our knowledge, no empirical or theoretical study has
been proposed in the literature to identify when and to what extent this
strategy should be used.
\end{abstract}

\section{Introduction}
\label{sc:introduction}
Constraint Satisfaction Problems (CSPs) and Constraint Optimisation Problems
(COPs) are defined on a set of variables representing problem entities.
Variables range on finite domains and are subject to a set of constraints
that define the feasible configurations of variable-value assignments. A COP
in addition has an objective function to be optimised. A solution to a CSP or
a COP is a variable-value assignment respecting all constraints, and
optimising the objective function if present. When being solved with
Constraint Programming, the solution process interleaves constraint
propagation and search.
%

A general way of building a search tree for solving CSPs and COPs is
called {\em labelling}. Labelling consists in selecting a variable and
assigning it a single value from its domain. The variable and value
selection are guided by heuristics. In particular, a value-selection
heuristic ranks values in such a way that the most promising value is 
selected first. Concerning value-selection heuristics, we consider the
following situations.

If the heuristic regards two or more values equally promising we say
the heuristic produces a tie, consisting of equally ranked domain
values. The definition of ties can be extended to the concept of
{\em heuristic equivalence} \cite{Gomes} that considers equivalent
all values that receive a rank within a given percentage from a value 
taken as reference.

A similar situation occurs when different domain value heuristics are
applied simultaneously. Often a problem is composed of
different aspects, for instance optimisation of profit, resource balance,
or feasibility of some problem constraints. For each of those aspects
a heuristic may be available. However, applying only one such
heuristic often does not lead to a globally satisfactory solution. The
goal is to combine these heuristics into one global domain value
heuristic. Many combinations are used in practise:
($i$) to follow the heuristic that is regarded most
important, and apply a different heuristic on values belonging to the
tie, ($ii$) to define a new heuristic (that might still contain ties)
as the (weighted) sum of the ranks that each heuristic assigns to a
domain value or ($iii$) rank the domain values through a {\em
  multi-criteria} heuristic. In this third case, a domain value has a
higher rank than another domain value if it has a higher rank with
respect to all heuristics. With respect to the multi-criteria
heuristic, some values may be incomparable. These incomparable values
together form a tie.

The two cases considered above describe the same situation: the
used heuristic(s) define(s) a {\em partial order} on the values'
ranks. In these cases, labelling chooses one of these values and
branches on it. In traditional tree search values are chosen according
to a deterministic rule, for instance lexicographic order. More
recently, randomisation has been applied to these choices, see
\cite{Gomes}. We propose a simple, yet effective method that improves
the efficiency of tree search in these situations: avoid making this
choice and postpone the branching decision.

Postponing branching decisions is practically used upon backtracking
in scheduling applications \cite{BookSched} when the {\em
  chronological heuristic} is chosen. We select the activity $A$ with
the smallest earliest start time $est$ and assign it to this
value. Upon backtracking, we postpone the decision for $A$ and go on
assigning a different activity. The motivation underlying this
postponement is that after a schedule that assigns activity $A$ to
$est$ is found (or the search has failed), it is unlikely that
assigning activity $A$ to $est+1$ would produce much better
results. Indeed, values $est$ and $est+1$ are equivalent (i.e., form a
tie) for the scheduling application.

We propose here to apply decision postponement systematically in case
of ties. Therefore, equivalent values are grouped together in a sub-domain
and a branching is performed on the whole sub-domain, while those
that are clearly ranked by the heuristic are still assigned singularly
to a variable. We call this method {\em partitioning}. 
Even this simple change can dramatically improve
the efficiency of the tree search, as we will see later. In addition,
partitioning has another important advantage: it enhances the bound
computation (in particular when used in conjunction with LDS), as
shown in \cite{LodiMilanoRousseauCP03} where a strategy using
partitioning is presented. Moreover, partitioning generates
sub-problems, to which any appealing search method may be applied,
speeding up the solution process. However, partitioning has also some
drawbacks. In particular, when constraint propagation heavily relies
on variable instantiation, partitioning may result in less
propagation. Nevertheless, when we apply a fast solution method to the
generated sub-problems, partitioning can still be favourable instead
of labelling.

Although domain partitioning and labelling have been already used for
solving CSPs and COPs, to our knowledge there is no theoretical and
practical study that indicates to practitioners when they should be
applied. In this paper we discuss the effect of domain partitioning to
search strategies that include depth-first search, limited discrepancy
search and variants.  

The outline of this paper is as follows. In
Section~\ref{sc:background} we define the concepts and the background
of our work. In Section~\ref{sc:theoretical} a theoretical comparison
of partitioning and labelling is given. This is followed by an
experimental comparison in Section~\ref{sc:experimental}. We conclude
with a discussion in Section~\ref{sc:discussion}.

\section{Background}\label{sc:background}
A constraint satisfaction problem (CSP) consists of a set of variables
$x_1, \dots, x_n$ with respective finite domains $D_1, \dots , D_n$,
and a set of constraints $C$ on these variables. A constraint
optimisation problem (COP) is a CSP together with an objective
function to be optimised.

We recall the concepts of labelling and partitioning, which are
standard and widely used in search strategies for solving CSPs and COPs.
During the search for a solution, a search tree is built by
subsequently taking branching decisions. A branching decision implies
first a variable selection. Then, {\em labelling} chooses a single
value and assigns the variable to that value. Upon backtracking
another value is assigned until no more values can be found in the
domain. Formally, for variable $x_i$, labelling generates the
assignments
\begin{displaymath}
x_i = d_{i_1} \vee x_i = d_{i_2} \vee \dots \vee x_i = d_{i_l}
\end{displaymath}
where $D_i = \{d_{i_1}, \dots, d_{i_l}\}$.

On the other hand {\em partitioning} is a technique that partitions
the domain of a variable and branches on the resulting sub-domains,
which may consist of only a single value.
A very simple example, widely used in CSPs, is to split a numerical
domain in two sets: the first containing values smaller or equal than
a given threshold $T$, the second containing values greater than $T$. For
example, if a variable $X$ ranges on a domain $\{1, \dots, 10\}$ the
partitioning can be $X \leq 5 \vee X >5$. This domain
can also be partitioned in different ways like $X \in \{4,5,6\} \vee X
\in \{1,3,7,8\} \vee X \in \{2,9,10\}$. Formally, for variable $x_i$,
partitioning generates the branching
\begin{displaymath}
x_i \in D_i^1 \vee x_i \in D_i^2 \vee \dots \vee x_i \in D_i^m
\end{displaymath}
where $D_i^1, \dots D_i^m$ is a partition of $D_i$. In this work, the
partition will be defined by the ties of the value-selection
heuristic, i.e. each $D_i^j$ consists of all values belonging to the
same tie.

Constructing a search tree via labelling leads to the appearance of
leaves only at depth $n$. Constructing a search tree via partitioning
leads to a sub-problem at depth $n$. If
all assigned sub-domains are single-valued, this sub-problem is a
leaf. Otherwise, the sub-problem must be searched again, through
labelling or partitioning. In this paper, we will always search the 
sub-problem via labelling. This means that the leaves of the search
tree appear at depth between $n$ and $2n$.

A search {\em strategy} defines the order in which the nodes of a
search tree are being traversed. We consider in this paper only
depth-first based search strategies. 
A depth-first based search strategy traverses the search
tree by going from a node to one of its successors, until it reaches a
leaf. Examples of depth-first based search strategies are
depth-first search (DFS), limited discrepancy search (LDS)
\cite{harvey95} and depth-bounded discrepancy search (DDS)
\cite{walsh_DBDS}.

%
%
A discrepancy (of a certain value) is a branching decision that is not
selected first by the domain value ordering heuristic. For LDS and
DDS, the cumulative discrepancy of a path from the root to a node may
not exceed a given limit. LDS gradually allows this limit to increase
during search. DDS follows LDS until a certain depth, but allows only
heuristic choices (discrepancy 0) below this depth.
The value of discrepancy of a branching decision is equal to the
number of preceding branching decisions at the current tree node. 
In case of labelling, the discrepancy increases with value 
1 for each domain value. For partitioning, the discrepancy value
increases with the number of domain values in each
sub-domain. However, below depth $n$, i.e. inside a sub-problem,
we say that no branching decision increases the discrepancy.

%

\section{Theoretical Comparison}\label{sc:theoretical}
This section shows, on a probabilistic basis, that partitioning is
more beneficial than labelling in case the (combined) heuristic
produces ties. In this section we do not consider constraint
propagation.

Similar to the analysis of LDS by Harvey and Ginsberg \cite{harvey95},
we introduce a probability that the heuristic makes a correct
choice. Let the search tree consist of good and bad nodes. A node is
called good if one of its successors is a (optimal) solution to the
CSP. Otherwise, the node is called bad. The heuristic probability is
the probability that at a good node, the heuristic selects a good node
first. Every following node selection has a similar probability of
being a good node. For simplicity,
Harvey and Ginsberg assume that this probability remains constant
throughout the search tree. To analyse DDS, Walsh \cite{walsh_DBDS}
introduces a similar probability, but explicitly assumes that it
increases with the depth. In both cases, binary search trees are 
considered, while our analysis is not restricted to binary trees.

The analysis of partitioning with respect to labelling should be based
on the sole fact that the heuristic produces ties. Hence, we may
assume that the heuristic probability remains constant throughout the
search tree. The heuristic probability is denoted by $p_i^d$,
corresponding to assigning value $d \in D_i$ to variable $x_i$.
Note that $\sum_{d \in D_i} p_i^d = 1$, and we explicitly assume
$p_i^{d_{i_j}} > p_i^{d_{i_k}}$ if the heuristic prefers $d_{i_j}$ over 
$d_{i_k}$. If the heuristic produces a tie for $x_i$, including values
$d_{i_j}$ and $d_{i_k}$, then $p_i^{d_{i_j}} = p_i^{d_{i_k}}$.

Let a search tree be defined by a certain variable ordering and domain
value heuristic. A leaf $l$ of the search tree consists of the
instantiation of all $n$ variables:
\begin{displaymath}
l = \{x_i = d_{i_j} \mid d_{i_j} \in D_i, i \in \{1, \dots, n\} \}.
\end{displaymath}
Thus, a leaf can either be a (optimal) solution or not. The
probability of a leaf $l$ being successful is
\begin{displaymath}
\prob(l) = \prod_{ \{x_i = d_{i_j} \} \in l} p_i^{d_{i_j}}.
\end{displaymath}

\begin{figure*}
\centerline{\includegraphics[width=.75\textwidth]{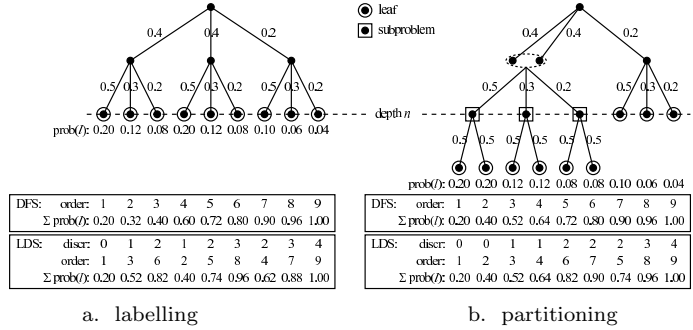}}
\begin{center}
\begin{footnotesize}
a.~~labelling \hspace{.28\textwidth} b.~~partitioning
\end{footnotesize}
\end{center}
\caption{Cumulative probability of success using DFS and LDS.} \label{fig:probs}
\end{figure*}

When we apply a certain search strategy to a tree defined by labelling
or by partitioning, leaves are visited in a different order. An example
of the probability distribution along the leaves of the different
search trees is given in Figure~\ref{fig:probs}. The trees correspond
to 2 variables, both having 3 domain values. The branches are ordered
from left to right following the heuristic's choice. The heuristic
probability of success for is shown for each branch. Note that the
heuristic produces a tie, consisting of two values, for the first
variable. Labelling follows the heuristic on single values, while 
partitioning groups together values in the tie. For DFS and LDS, the
order in which the leaves are visited is given, together with the
cumulative probability of success. Note that for every
leaf, partitioning always has a higher (or equal) cumulative
probability of success than labelling. This will be formalised in
Theorem~\ref{th:compare}.
Note also that in a sub-problem generated by partitioning all leaves
have the same probability of success. This property follows
immediately from the construction of the sub-problems. As a
consequence, any search strategy applied to this sub-problem will be
equally likely to be successful. In practise, we will therefore use
DFS to solve the sub-problems.

\begin{theorem} \label{th:compare}
For a fixed variable ordering and a domain value ordering heuristic,
let $T_{\lab}$ be the search tree defined by labelling, and let 
$T_{\prt}$ be the search tree defined by partitioning, grouping 
together ties. Let the set of the first $k$ leaf nodes visited by 
labelling and partitioning be denoted by $L_{\lab}^k$ and $L_{\prt}^k$ 
respectively. If $T_{\lab}$ and $T_{\prt}$ are traversed using the 
same depth-first based search strategy then
\begin{equation}\label{eq:part_label}
\sum_{l \in L_{\prt}^k} \prob(l) \geq
\sum_{l \in L_{\lab}^k} \prob(l).
\end{equation}
\end{theorem}
\begin{proof}
For $k = 1$, (\ref{eq:part_label}) obviously holds. Let $k$ increase
until labelling and partitioning visit a leaf with a different
probability of success, say $l_k^{\lab}$ and $l_k^{\prt}$
respectively. If such leaves do not exist, (\ref{eq:part_label}) holds
with equality for all $k$.

Assume next that such leaves do exist, and let $l_k^{\lab}$ and 
$l_k^{\prt}$ be the first leaves with a different probability of success. 
As the leafs are different, there is at least one different branching decision
between the two. The only possibility for this different branching
decision is that we have encountered a tie, because partitioning and
labelling both follow the same depth-first based search strategy. This
tie made partitioning create a sub-problem $S$, with $l_k^{\prt}
\in S$, and $l_k^{\lab} \notin S$. If labelling made a
branching decision different from partitioning, with a higher
probability of being successful, then partitioning would have made the
same decision. Namely, partitioning and labelling follow the same
strategy, and the heuristic prefers values with a higher
probability. So it must be that a different branching decision made by
labelling has a smaller or equal probability of being successful with
respect to the corresponding decision made by partitioning. However,
as we have assumed that $\prob(l_k^{\prt}) \neq \prob(l_k^{\lab})$,
there must be at least one different branching decision made by
labelling, that has a strictly smaller probability of being
successful. Thus for the current $k$, (\ref{eq:part_label}) holds, and
the inequality is strict.

As we let $k$ increase further, partitioning will visit first all
leaves inside $S$, and then continue with $l_k^{\lab}$. On the other
hand, labelling will visit leaves $l$ that are either in $S$ or
not, all with $\prob(l) \leq \prob(l_k^{\prt})$. However, as
partitioning follows the same search strategy as labelling,
partitioning will either visit a leaf of a sub-problem, or a leaf that
labelling has already visited (possibly simultaneously). In both
cases, $\sum_{l \in L_{\prt}^k} \prob(l) \geq \sum_{l \in L_{\lab}^k}
\prob(l)$.
\end{proof}

\begin{figure*}
\centerline{\includegraphics[width=.49\textwidth]{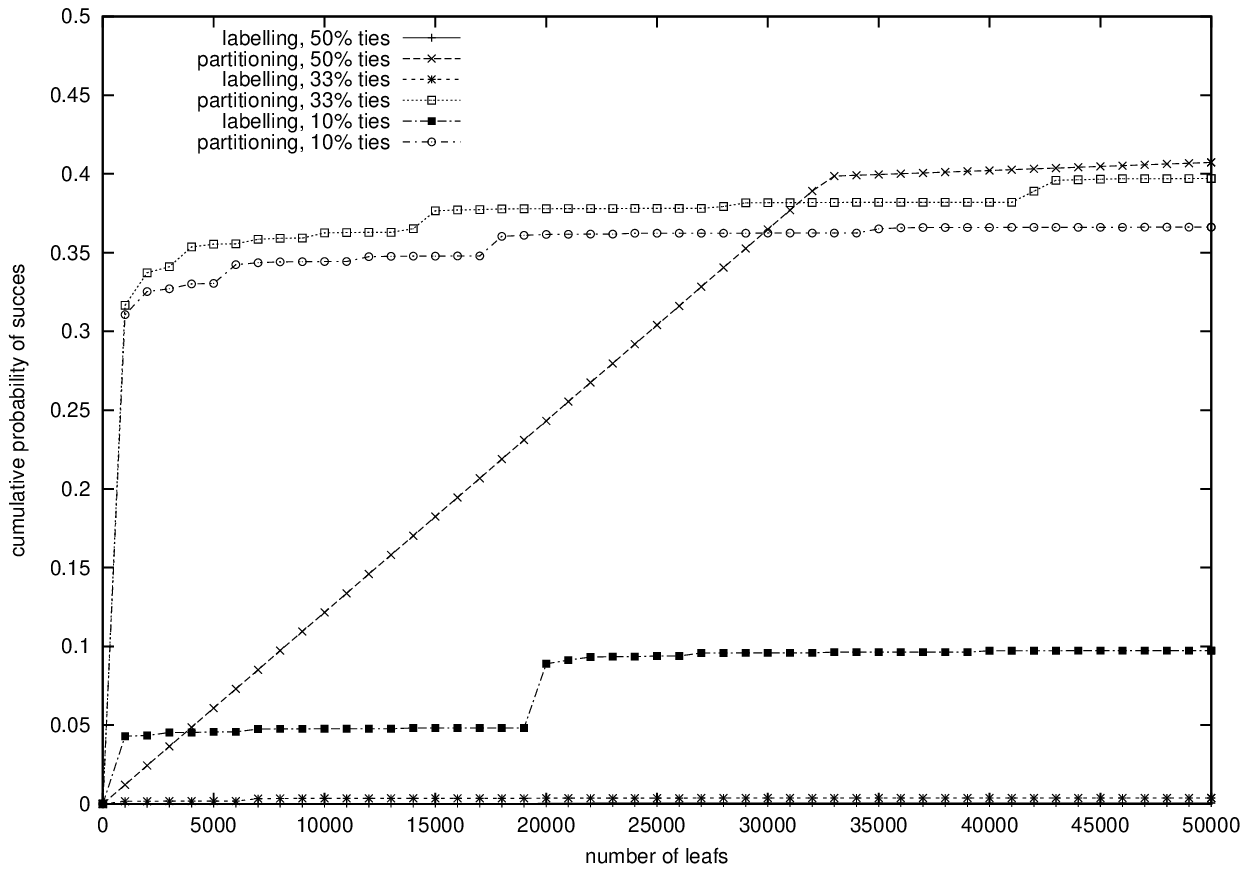}
\hfill
\includegraphics[width=.49\textwidth]{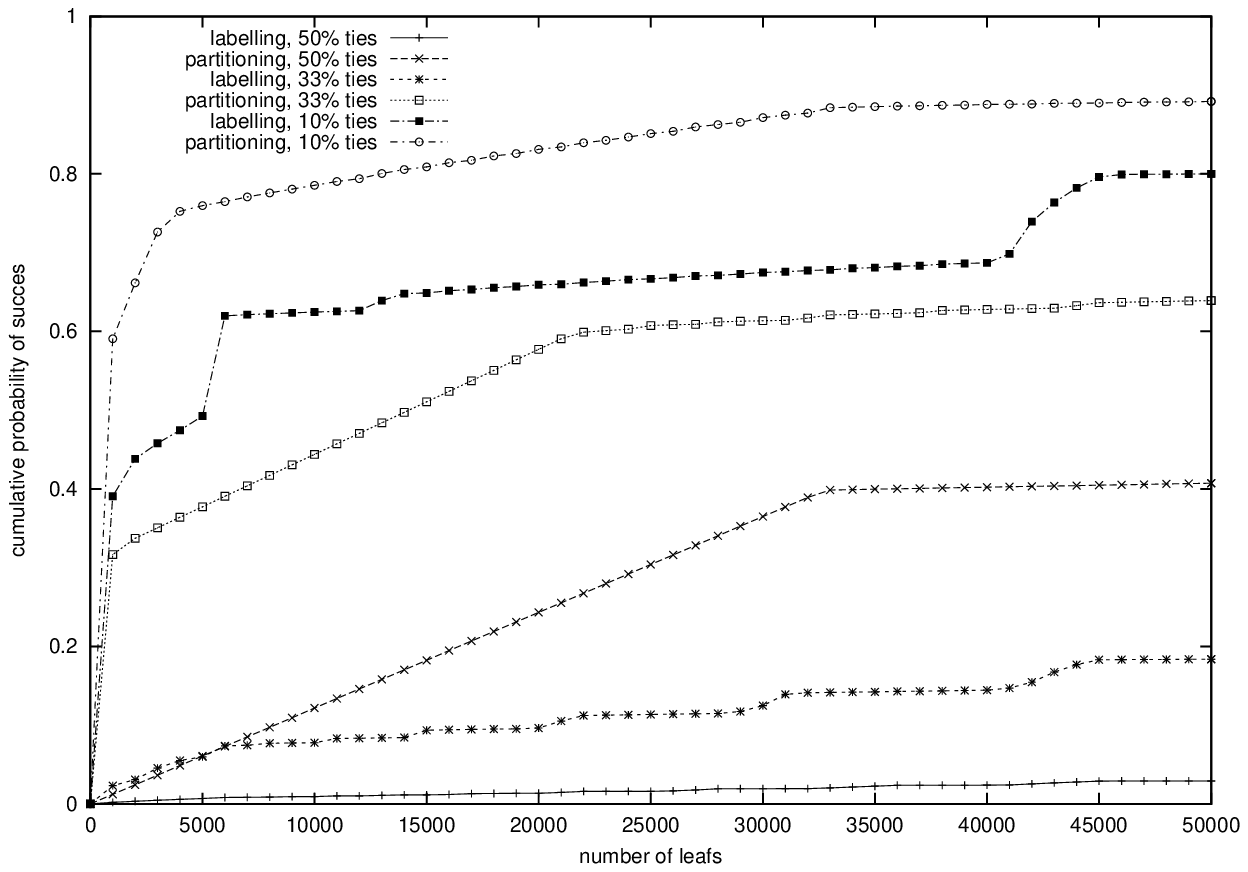}}
\begin{center}
\begin{footnotesize}
a.~~using DFS \hspace{.4\textwidth} b.~~using LDS
\end{footnotesize}
\end{center}
\caption{Partitioning versus labelling on search trees of
  depth 30 and branch-width 3.} \label{fig:versus}
\end{figure*}

Next we measure the effect that the number of ties has on the
performance of partitioning with respect to labelling. For this reason,
we vary the number of ties in a fixed search tree of depth 30. A
branch-width of 3 will be used in all cases, as this allows ties, and a
larger branch-width would make it impractical to measure effectively
the performance of labelling. Depending on the occurrence of a tie, the
heuristic probability $p_i$ of branch $i$ will be chosen either
\begin{itemize}
\item[] $p_1 = 0.95$, $p_2 = 0.04$, $p_3 = 0.01$ (no tie), or
\item[] $p_1 = 0.495$, $p_2 = 0.495$, $p_3 = 0.01$ (tie).
\end{itemize}
Our method assumes a fixed variable ordering in the search tree, and
uniformly distributes the ties among them. This is reasonable, since
in practise ties can appear unexpectedly. We have investigated the
appearance of 10\%, 33\% and 50\% ties out of the $n$ branching
decisions that lead to a leaf. In Figure~\ref{fig:versus}.a and b., we
report the cumulative probability of success for labelling and
partitioning using DFS and LDS until 50000 leaves. Note that in
Figure~\ref{fig:versus}.a the graphs for labelling with 33\% and 50\%
ties almost coincide along the x-axis. The figures show
that in the presence of ties partitioning may be much more beneficial 
than labelling, i.e. the strict gap in (\ref{eq:part_label}) can be 
very large.

\section{Experimental Comparison} \label{sc:experimental}
This section presents computational results of two applications for which
we have compared partitioning and labelling. The first is the Travelling
Salesman Problem (TSP), the second the Partial Latin Square Completion
Problem (PLSCP). We first explain the reason why we chose these two
problems among a set of problems considered to test the methods. The
TSP is an optimisation problem where the propagation is 
quite poor and the heuristic used is very informative but produces (not
very large indeed) ties. Instead, PLSCP is a constraint satisfaction
problem whose model contains many {\em all\-different} constraints whose
filtering algorithm is particularly effective. The heuristic used is quite
good and sometimes produces ties. Therefore, the two problems have opposite
structure and characteristics. For the TSP partitioning is very suitable
since the only drawback of the method, i.e., the decreased effect of
propagation, does not play any role. On the contrary, the PLSCP is a
problem whose characteristics are not suitable for the partitioning.
Therefore, we will point out also the weakness of the method. For each
application we state the problem, define the applied heuristic and report
the computational results. For both problems we apply LDS as search
strategy.

The applications are implemented on a Pentium 1Ghz with 256 MB
RAM, using ILOG Solver~5.1 \cite{ilog51} and Cplex~7.1 \cite{cplex71}.

\subsection{Travelling Salesman Problem}\label{ssc:tsp}
The travelling salesman problem (TSP) is a traditional NP-hard
combinatorial optimisation problem. Given a set of cities with
distances (costs) between them, the problem is to find a closed tour
of minimal length visiting each city exactly once.

For the TSP, we have used a constraint programming model and a heuristic
similar to~\cite{milano_hoeve02} based on reduced costs. Sub-problems are
being solved using DFS, since all leaves can be considered to have equal
probability of being successful.

To compare labelling and partitioning fairly, we stop the search as soon as
an optimal solution has been found. For the considered instances, the
optimal values are known in advance. The proof of optimality should not be
taken into account, because it is not directly related to the probability
of a branch being successful.

The results of our comparison are presented in Table~\ref{tb:tsp}.
The instances are taken from TSPLIB \cite{TSPLIB} and represent
symmetric TSPs. For labelling and partitioning, the table shows the
time and the number of fails (backtracks) needed to find an
optimum. For labelling, the discrepancy of the leaf node that
represents the optimum is given. For partitioning, the discrepancy of
the sub-problem that contains the optimum is reported.

For all instances but one, partitioning performs much better
than labelling. Both the number of fails and the computation time are
substantially less for partitioning. Observe that for the instance
`dantzig42' labelling needs less fails than partitioning, but uses
more time. This is because partitioning solves the sub-problems using
DFS. Partitioning can visit almost three times more nodes in less
time, because it lacks the LDS overhead inside the sub-problems.

\begin{table}
\begin{center}
\begin{footnotesize}
\begin{tabular}{|l|rrc|rrc|} \hline \hline
 & \multicolumn{3}{|c|}{\bf labelling} & \multicolumn{3}{|c|}{\bf partitioning} \\
instance & time (s) & fails & discr & time (s) & fails & discr  \\ \hline
gr17 & 0.08 & 36 & 2 & 0.02 & 3 & 0 \\
gr21 & 0.16 & 52 & 3 & 0.01 & 1 & 0 \\
gr24 & 0.49 & 330 & 5 & 0.01 & 4 & 0 \\
fri26 & 0.16 & 82 & 2 & 0.01 & 0 & 0 \\
bayg29 & 8.06 & 4412 & 8 & 0.07 & 82 & 1 \\
bays29 & 2.31 & 1274 & 5 & 0.07 & 43 & 1 \\
dantzig42 & 0.98 & 485 & 1 & 0.79 & 1317 & 1 \\
swiss42 & 6.51 & 2028 & 4 & 0.08 & 15 & 0 \\
hk48 & 190.96 & 35971 & 11 & 0.23 & 175 & 1 \\
brazil58 & N.A. & N.A. & N.A. & 0.72 & 770 & 1 \\ \hline \hline
\end{tabular}\\ \vspace{1ex}
N.A. means `not applicable' due to time limit (900 s).
\end{footnotesize}
\end{center}
\caption{Results for finding optima of TSP instances (not proving
  optimality).}\label{tb:tsp}
\end{table}

\subsection{Partial Latin Square Completion Problem}\label{ssc:pls}
The Partial Latin Square Completion Problem (PLSCP) is a well known
NP-complete combinatorial satisfaction problem. A Latin square is an
$n \times n$ square in which each row and each column is a permutation
of the numbers $\{1, \dots, n\}$. A partial Latin square is a
partially pre-assigned square. The PLSCP is the problem of extending a
partial Latin square to a feasible (completely filled) Latin square.

\begin{table}
\begin{center}
\begin{footnotesize}
\begin{tabular}{|l|rp{.7cm}c|rp{.7cm}c|} \hline \hline
 & \multicolumn{3}{|c|}{\bf labelling} & \multicolumn{3}{|c|}{\bf partitioning} \\
instance & time (s) & fails & discr & time (s) & fails & discr  \\ \hline
b.o25.h238 & 2.36 & 668 & 5 & 1.09 & 746 & 5 \\
b.o25.h239 & 0.49 & 15 & 1 & 0.42 & 2 & 1 \\
b.o25.h240 & 1.17 & 179 & 4 & 0.86 & 893 & 4 \\
b.o25.h241 & 3.31 & 772 & 3 & 4.70 & 3123 & 4 \\
b.o25.h242 & 2.41 & 537 & 3 & 1.80 & 1753 & 4 \\
b.o25.h243 & 4.06 & 1082 & 4 & 3.96 & 2542 & 4 \\
b.o25.h244 & 1.33 & 214 & 3 & 2.99 & 2072 & 4 \\
b.o25.h245 & 9.40 & 2308 & 6 & 10.66 & 12906 & 7 \\
b.o25.h246 & 2.01 & 401 & 5 & 2.22 & 1029 & 4 \\
b.o25.h247 & 258.91 & 69105 & 6 & 11.66 & 5727 & 4 \\
b.o25.h248 & 33.65 & 6969 & 5 & 0.68 & 125 & 2 \\
b.o25.h249 & 212.76 & 60543 & 11 & 101.46 & 85533 & 8 \\
b.o25.h250 & 2.45 & 338 & 2 & 0.83 & 687 & 3 \\
u.o30.h328 & 273.53 & 32538 & 4 & 82 & 14102 & 3 \\
u.o30.h330 & 21.79 & 2756 & 3 & 25.15 & 5019 & 3 \\
u.o30.h332 & 235.40 & 30033 & 5 & 56.94 & 9609 & 3 \\
u.o30.h334 & 4.18 & 256 & 2 & 6.09 & 843 & 2 \\
u.o30.h336 & 1.73 & 69 & 2 & 0.76 & 12 & 1 \\
u.o30.h338 & 49.17 & 5069 & 3 & 29.41 & 8026 & 3 \\
u.o30.h340 & 1.68 & 91 & 2 & 0.81 & 66 & 2 \\
u.o30.h342 & 28.40 & 3152 & 3 & 5.41 & 600 & 2 \\
u.o30.h344 & 9.05 & 605 & 2 & 8.35 & 1103 & 2 \\
u.o30.h346 & 2.15 & 101 & 2 & 3.76 & 482 & 2 \\
u.o30.h348 & 43.80 & 2658 & 2 & 32.86 & 2729 & 2 \\
u.o30.h350 & 1.16 & 46 & 1 & 0.80 & 12 & 1 \\
u.o30.h352 & 5.10 & 288 & 2 & 0.95 & 32 & 1 \\ \hline
sum & 1211.45 & 220793 & 91 & 396.62 & 159773 & 81 \\
mean & 46.59 & 8492.04 & 3.50 & 15.25 & 6145.12 & 3.12 \\ \hline \hline
\end{tabular}
\end{footnotesize}
\end{center}
\caption{Results for PLS completion problems.}\label{tb:latin}
\end{table}

The constraint programming model is straightforward, using
{\em all\-different} constraints on the rows and the columns, with
maximal propagation. The maximal {\em alldifferent} propagation
(achieving hyper-arc consistency \cite{regin94}) is of great
importance for solving the PLSCP. With less powerful propagation, the
considered instances are practically unsolvable.

As heuristic we have used a simple first-fail principle for the
values, i.e. values that are most constrained are to be considered
first. Therefore the rank of a value is taken equal to the number of
the value's occurrences in the partial Latin square, and a higher rank
is regarded better. Hence, labelling selects the value with the highest
rank, and uses lexicographic ordering in case of ties. Partitioning
selects the sub-domain consisting of all values having the highest
rank. The sub-problems are again being solved using DFS. For both
labelling and partitioning, constraint propagation is
applied throughout the whole search tree.

In Table~\ref{tb:latin} we report the performance of labelling and
partitioning on a set of partial Latin square completion problems. It 
follows the same format as Table~\ref{tb:tsp}. The
instances are generated with the PLS-generator
\cite{gomes_color02}. Following remarks made in \cite{gomes_color02},
our generated instances are such that they are difficult to solve,
i.e. they appear in the transition phase of the problem. The instances
`b.o25.h$m$' are balanced $25 \times 25$ partial Latin squares, with
$m$ unfilled entries (around 38\%). Instances `u.o30.h$m$' are
unbalanced $30 \times 30$ partial Latin squares, with $m$ unfilled
entries (around 38\%).

Although partitioning performs much better than labelling on
average, the results are not homogeneous. For some instances labelling has
better performances w.r.t. partitioning. This can be explained by the
pruning power of the {\em alldifferent} constraint. Since partitioning
branches on sub-domains of cardinality larger than one, the {\em
alldifferent} constraint will remove less inconsistent values compared to
branching on single values, as is the case with labelling. Using
partitioning, such values will only be removed inside the sub-problems.
However, even in instances where partitioning is less effective, the
difference between the two strategies is not so high, while on many
instances partitioning is much more effective.

As was already mentioned in Section~\ref{ssc:tsp}, partitioning effectively
applies DFS inside the sub-problems. For a number of instances, partitioning
finds a solution earlier than labelling, although making use of a higher
number of fails.

\section{Discussion}\label{sc:discussion}
We have seen both theoretically and experimentally that partitioning
is to be preferred over labelling, when some domain values are
incomparable with respect to one or more heuristics. There are several
additional benefits to partitioning, of which we would like to mention
two. Thereafter we discuss various drawbacks of partitioning.

The sub-problems that are created by partitioning may be subject to
any applicable search method. In particular, when the
sub-problems are large, one could apply a local search method. Another
possibility is to apply a (mixed-integer) (non)linear programming
solver. This allows the user to effectively combine several solution
methods to solve the problem.

For COPs, proving optimality is often more difficult than finding a
good solution. Partitioning can sometimes be useful to prove
optimality earlier. In \cite{milano_hoeve02}, partitioning is applied
to a domain value ordering heuristic based on reduced costs, together
with LDS. For that particular case, the partitioning scheme allows a
very effective bound computation.

On the other hand, branching on sub-domains instead of single values
decreases the effect of constraint propagation. This is a serious
drawback of partitioning, as we have seen in Section~\ref{ssc:pls}.
It also affects the bound computation of COPs. As was suggested in
\cite{LodiMilanoRousseauCP03}, `additive bounding' procedures may be
helpful in this case.

Finally, we have only considered partitioning on depth-first based
search strategies. We are currently investigating the possibility
to effectively apply partitioning to breadth-first based search
strategies as well.

\end{document}